\documentclass[letterpaper]{article} 
\usepackage{aaai25}  
\usepackage{times}  
\usepackage{helvet}  
\usepackage{courier}  
\usepackage[hyphens]{url}  
\usepackage{graphicx} 
\usepackage{natbib}  
\usepackage{caption} 
\usepackage{algorithm}
\usepackage{algorithmic}

\usepackage{newfloat}
\usepackage{listings}
\DeclareCaptionStyle{ruled}{labelfont=normalfont,labelsep=colon,strut=off} 
\floatstyle{ruled}
\newfloat{listing}{tb}{lst}{}
\floatname{listing}{Listing}
\title{Enhancing Uncertainty Modeling with Semantic Graph for Hallucination Detection}
\author{
    Kedi Chen\textsuperscript{\rm 1}\equalcontrib\thanks{This work was done during the internship at Xiaohongshu Inc.}, Qin Chen\textsuperscript{\rm 1}\equalcontrib\thanks{Corresponding author.}, Jie Zhou\textsuperscript{\rm 1}, Xinqi Tao\textsuperscript{\rm 2}, Bowen Ding\textsuperscript{\rm 2}, Jingwen Xie\textsuperscript{\rm 2}, Mingchen Xie\textsuperscript{\rm 2}, Peilong Li\textsuperscript{\rm 2}, Feng Zheng\textsuperscript{\rm 2}, Liang He\textsuperscript{\rm 1}
}
\affiliations{
    \textsuperscript{\rm 1}East China Normal University \\ \textsuperscript{\rm 2}Xiaohongshu Inc.\\
    kdchen@stu.ecnu.edu.cn \{qchen, jzhou, lhe\}@cs.ecnu.edu.cn \\
    \{yifan5, faming, qingliang, shenzong, liaofan, yemu\}@xiaohongshu.com

}

\usepackage{bibentry}

\usepackage{multirow}
\usepackage{booktabs}
\usepackage{colortbl}
\usepackage[table]{xcolor}
\usepackage{mathrsfs}
\usepackage{amsmath}
\usepackage{tikz}

\begin{document}

\maketitle

\begin{abstract}
Large Language Models (LLMs) are prone to hallucination with non-factual or unfaithful statements, which undermines the applications in real-world scenarios. 
Recent researches focus on uncertainty-based hallucination detection, which utilizes the output probability of LLMs for uncertainty calculation and does not rely on external knowledge or frequent sampling from LLMs. 
Whereas, most approaches merely consider the uncertainty of each independent token, while the intricate semantic relations among tokens and sentences are not well studied, which limits the detection of hallucination that spans over multiple tokens and sentences in the passage. 
In this paper, we propose a method to enhance uncertainty modeling with semantic graph for hallucination detection. 
Specifically, we first construct a semantic graph that well captures the relations among entity tokens and sentences. 
Then, we incorporate the relations between two entities for uncertainty propagation to enhance sentence-level hallucination detection. 
Given that hallucination occurs due to the conflict between sentences, we further present a graph-based uncertainty calibration method that integrates the contradiction probability of the sentence with its neighbors in the semantic graph for uncertainty calculation. 
Extensive experiments on two datasets show the great advantages of our proposed approach. 
In particular, we obtain substantial improvements with 19.78\% in passage-level hallucination detection.

\end{abstract}


\section{Introduction}
Large Language Models (LLMs) \citep{DBLP:journals/corr/abs-2303-18223}, with large-scale parameters and advanced training methods, achieve excellent performance in many downstream tasks of natural language processing (NLP) \citep{DBLP:journals/corr/abs-2401-04637,DBLP:conf/coling/Chen0C0024,DBLP:journals/csur/LaiN24,DBLP:journals/corr/abs-2401-13601}.
Despite the many benefits of large language models, hallucination remains an issue that cannot be ignored.
Hallucination indicates that some non-factual or untruthful contents are generated \citep{DBLP:journals/corr/abs-2310-07521}.
Therefore, hallucination detection is critically an essential task, which provides a preliminary review of the contents generated by large language models, reducing their potential harm in real-world scenarios \citep{DBLP:journals/corr/abs-2402-02315,DBLP:journals/corr/abs-2306-16092,DBLP:journals/bjet/YanSZLMCLJG24}, such as education, economics, science, and so on. 

\begin{figure}[t]
\centering
\includegraphics[width=1.0\columnwidth]{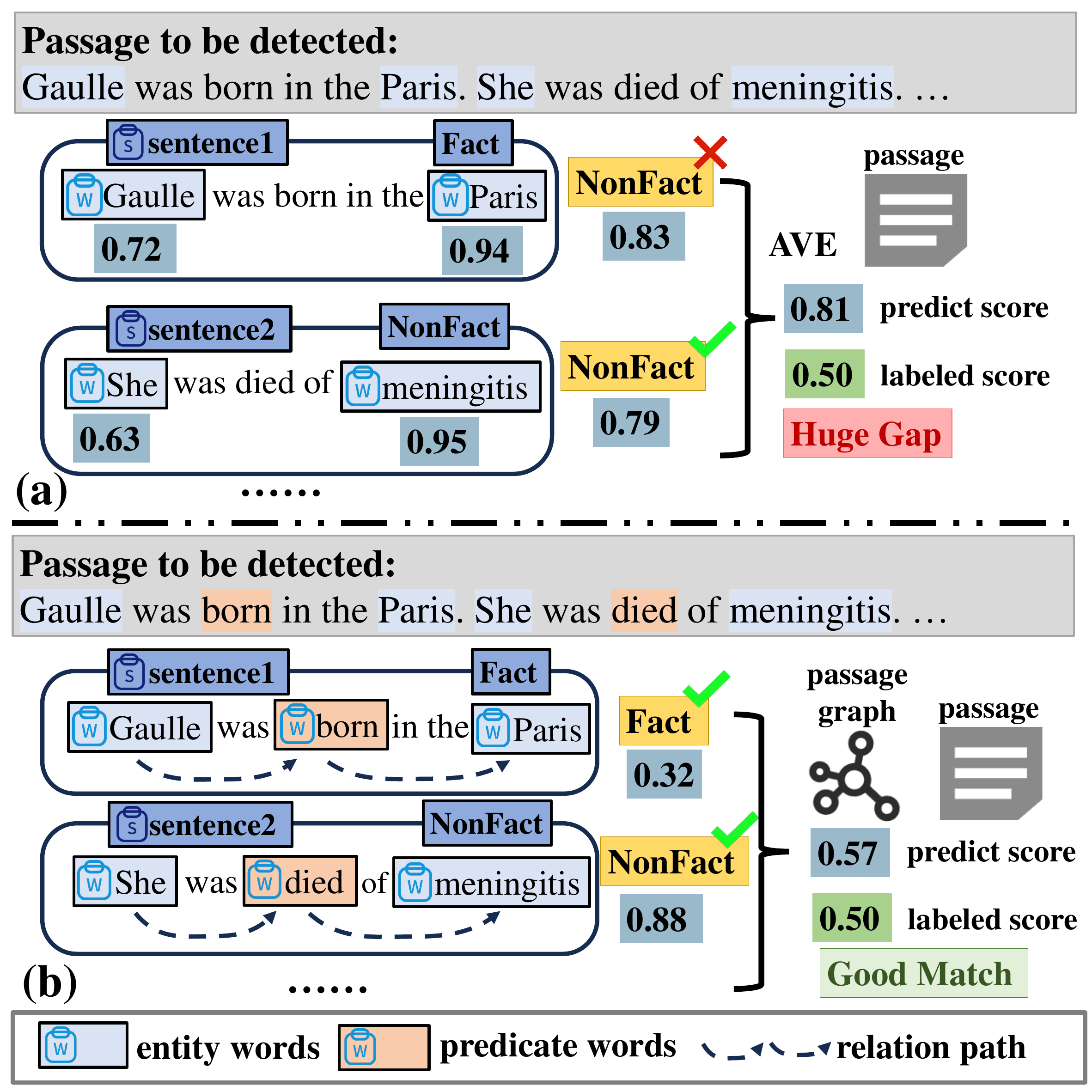} 
\caption{(a) Previous works only concern independent tokens and use their average scores as the metrics, resulting in errors in sentence and passage-level detection. (b) Our method captures more complex semantic dependencies with a semantic graph for uncertainty modeling, such as the relations between entities, and the relations with neighbor sentences in the passage-level semantic graph.}
\label{fig1}
\end{figure}


Current hallucination detection methods can be roughly divided into three categories.
(i) Retrieval-based method \citep{DBLP:conf/emnlp/WangYHZH23,DBLP:conf/acl/ZhangMA0CZCXZC023} usually retrieve evidence from external resources for fact verification \citep{DBLP:conf/aaai/0011LH024}. 
This approach exceedingly depends on the quality of external resources, which is not always available. 
In addition, it needs various validation steps towards the retrieved knowledge, which are complicated and inefficient. 
(ii) Sampling-based method frequently samples responses from LLMs for consistency verification, which consumes substantial computational resources \citep{DBLP:conf/emnlp/ManakulLG23,DBLP:conf/emnlp/ZhangLDMS23}. 
(iii) The uncertainty-based method is a good alternative to resolve the above problems \citep{DBLP:conf/emnlp/GiulianelliBAFP23,DBLP:journals/corr/abs-2306-13063}. 
It leverages LLMs to output the probability of each token in the text to be detected and then computes a hallucination score with uncertainty-based metrics. 
Given that this method requires the models to perform inference only once, it is relatively efficient and thus attracts increasing interest from researchers.

Nevertheless, several challenges persist in uncertainty-based methods for hallucination detection (Figure~\ref{fig1}).
\textbf{First,} most methods focus on modeling the uncertainty of each independent token, while the complex dependency among tokens within the sentence is not well explored. 
Recent methods \cite{DBLP:conf/emnlp/ZhangQGDZZZWF23} tend to propagate the uncertainties of all previous tokens to the subsequent ones for uncertainty calculation. 
However, not all tokens are semantically related, and this propagation leads to uncertainty overestimation as shown in Figure~\ref{fig1}. 
\textbf{Second,} passage-level uncertainty is not well studied. 
Previous methods usually average the uncertainty score of each sentence \citep{DBLP:conf/emnlp/ManakulLG23,DBLP:conf/emnlp/ZhangQGDZZZWF23}, while neglecting the intricate relations such as the semantic conflicts among sentences in the whole passage.


To resolve the above two challenges, we propose an approach to enhance uncertainty modeling with semantic graph for hallucination detection. Specifically, we first perform Abstract Meaning Representation (AMR) \citep{xu-etal-2023-learning-friend} based parsing for each sentence, and obtain a passage-level AMR graph by coreference resolution and entity linking between sentences, which well captures the semantic dependency relations among the entity tokens and the sentences for hallucination detection. 
Then, we present a relation-based propagation method, which propagates the uncertainty from one entity to the other along the relation path in the semantic graph to enhance sentence-level hallucination detection as shown in Figure~\ref{fig1}. 
Regarding passage-level hallucination detection, we further integrate the relations between the sentence and its neighbors in the graph for uncertainty calibration via the natural language inference (NLI) \citep{DBLP:conf/acl/ZhengZ23} technique. 

We perform experiments on two datasets, namely the well-known WikiBio \citep{DBLP:conf/emnlp/ManakulLG23} and our constructed NoteSum.
The results show the great superiority of our approach in both sentence-level and passage-level hallucination detection.

The main contributions can be summarized as follows:
\begin{itemize}
\item To the best of our knowledge, it is the first attempt to explore the potential of semantic graph to capture the complex relations among the tokens and the sentences for hallucination detection.
\item We present two novel methods, namely relation-based uncertainty propagation and graph-based uncertainty calibration, which shed light on how to integrate the structured semantic graph with the uncertainty computation framework.
\item We conduct elaborate analyses of the experimental results on two benchmark datasets, and provide a better understanding of the effectiveness of our approach\footnote{https://github.com/141forever/UncerSema4HalluDetec}.
\end{itemize}

\section{Related Work}
\subsection{Hallucination in Language Models}
Hallucination reflects that language models generate some nonsensical or untruthful contents \citep{DBLP:journals/corr/abs-2310-07521} in many downstream NLP tasks, such as the question and answer task \citep{DBLP:conf/emnlp/NaszadiMM23}, the multi-turn dialogue task \citep{DBLP:journals/corr/abs-2403-00896} and the text summarization task \citep{DBLP:conf/emnlp/KryscinskiMXS20}, etc.
Hallucination in NLP can be categorized into two main classes: factuality hallucination and faithfulness hallucination \citep{DBLP:journals/corr/abs-2311-05232}.
The former one reveals the generated contents contain factual errors against real life, while the latter demonstrates the issues of inconsistency or irrelevance in the text.

\subsection{Hallucination Detection}
Before the era of LLMs, researchers normally train a discriminating model to judge whether hallucination exists \citep{DBLP:conf/emnlp/ZhaoND23}. 
This approach relies too heavily on the training data and can reduce the models’ generalization ability.
With the development of NLP technology, current hallucination detection methods can be roughly divided into three categories.

\textbf{Retrieval-based method} \citep{DBLP:conf/emnlp/WangYHZH23,DBLP:conf/acl/ZhangMA0CZCXZC023} utilizes the retrieval-augmented generation technique \citep{DBLP:conf/aaai/0011LH024} for extra knowledge \citep{DBLP:conf/emnlp/ChoiF0S23} or information to help detection \citep{DBLP:journals/corr/abs-2307-03987,DBLP:journals/corr/abs-2403-00896,siino-2024-brainllama}.
This approach exceedingly depends on the quality of information sources, necessitating complicated validation steps \citep{DBLP:conf/sigir/YeLYC0024,DBLP:conf/sigir/DongLA0L0WYM24} towards the retrieved knowledge. 
Not to mention that not all information is available easily. 
On the contrary, we propose an efficient reference-free method.

\textbf{Sampling-based method} rewrites the contents under detection, measuring the consistency and coherence \citep{DBLP:conf/acl/MalkinWJ22,DBLP:conf/aaai/ShengZJK24} between them to acquire a hallucination score \citep{DBLP:conf/emnlp/ManakulLG23,DBLP:conf/emnlp/ZhangLDMS23,DBLP:journals/corr/abs-2310-17918,mündler2024selfcontradictoryhallucinationslargelanguage}. 
However, this strategy frequently invokes LLMs for rewriting, consuming substantial computational resources.
Our method needs one LLM to infer only once, thereby greatly saving the response time.

\textbf{Uncertainty-based method} applies proxy-based LLMs to output the probability of each token in contents to be detected and then estimates a hallucination score with uncertainty-based metrics \citep{DBLP:journals/corr/abs-2307-10236,DBLP:conf/emnlp/ChenDBQWCW23,wang2023uncertaintyawareparameterefficientselftrainingsemisupervised,petersen2024uncertaintyquantificationstabledistribution,xiong2024can}.
\citet{DBLP:conf/emnlp/ManakulLG23} regards the degree of hallucination as being negatively correlated with the probability.
\citet{DBLP:conf/emnlp/ZhangQGDZZZWF23} refutes this view, but there arises a co-occurrence bias \citep{DBLP:journals/corr/abs-2307-15252}.
Due to a lack of detailed exploration of various dependencies, our method systematically constructs the relationships among the entity tokens and the sentences.

\begin{figure*}[!t]
\centering
\includegraphics[width=1.0\textwidth]{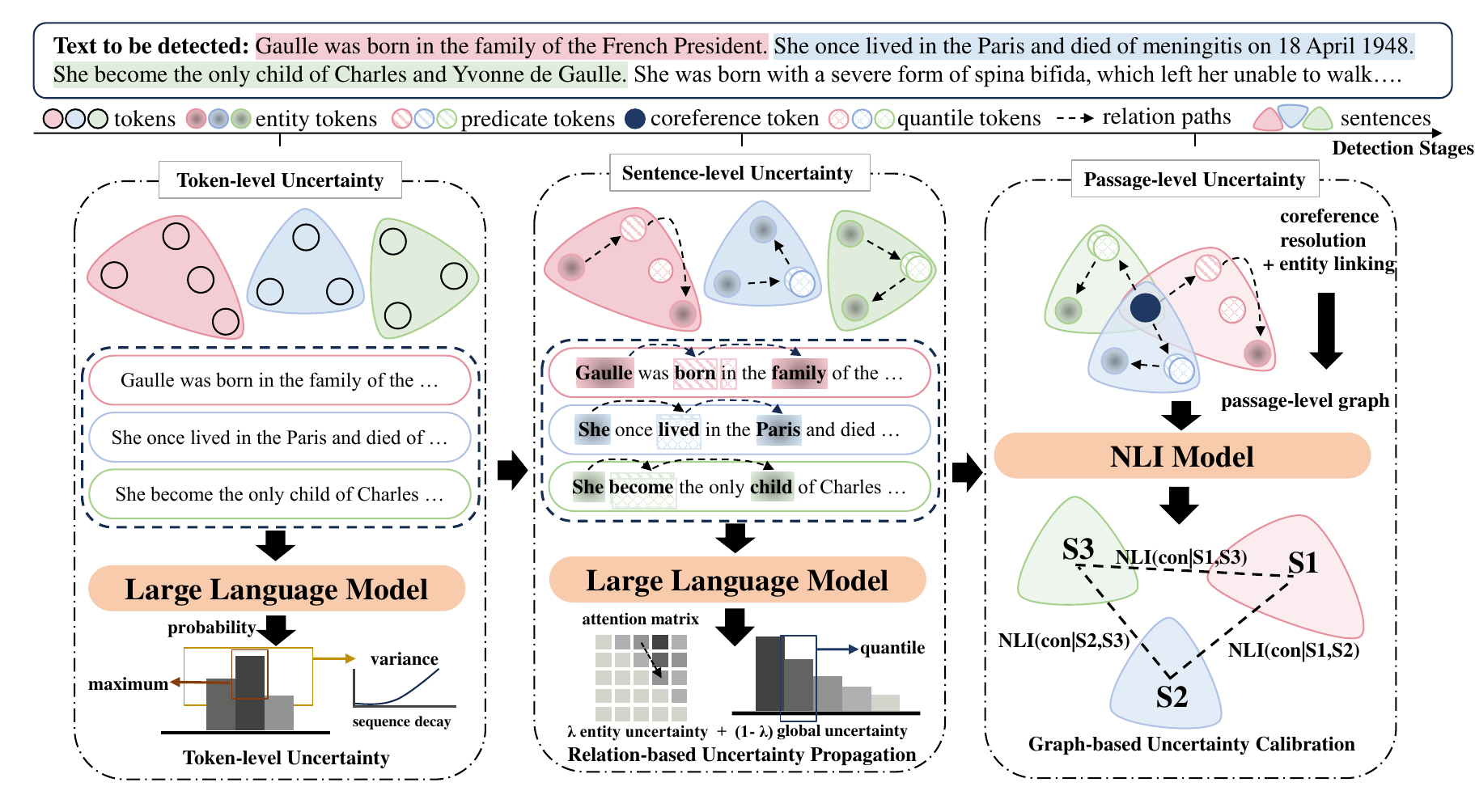} 
\caption{The overview of our approach for hallucination detection. For token-level uncertainty, we integrate the maximum and variance of the probabilities, along with a sequence decay term. Regarding to sentence-level uncertainty, we interpolate the sum of entity uncertainty through relation-based propagation and global uncertainty via quantile. Finally, we incorporate the relations of neighbor sentences in the semantic graph with graph-based uncertainty calibration for passage-level uncertainty. }
\label{fig:overview}
\end{figure*}

\section{Our Approach}
The framework of our proposed approach is illustrated in Figure~\ref{fig:overview}. 
Specifically, inspired by the findings that hallucination accumulates as the sequence length increases, we integrate the distribution statistics of LLM-based conditional probability with sequence decay for token-level uncertainty calculation. 
Considering much hallucination is induced by the entities and relations in the sentence and passage, we further construct a semantic graph for sentence-level and passage-level uncertainty calculation. 
Regarding sentence-level uncertainty, it well captures the semantic relations between entities for hallucination propagation and calculation. 
In particular, the uncertainty of an entity propagates to the related entity along the dependent relations. 
For passage-level uncertainty, we incorporate the neighbors of each sentence in the semantic graph for uncertainty calibration and summation. 
The details are denoted in the following.



\paragraph{Semantic Graph Construction.}
To better model the uncertainties of entities with long-range dependency that span over the text, we first perform AMR \citep{xu-etal-2023-learning-friend} parsing for each sentence, and gain a sentence-level graph where each node is an entity and the edge represents the dependent semantic relation. 
Compared to traditional dependency parsing, AMR parsing is more logical and less vulnerable to syntactic representation or word order variations. 
Therefore, we employ AMR to model the inter-dependency between the entities in the sentence. 
Furthermore, noting that passage-level hallucination usually occurs when two sentences contradict each other, we further link sentence-level AMR graphs together by the intricate relations (e.g., entity linking and coreference) among sentences. 
Finally, a large AMR graph corresponding to the passage is acquired.

Formally, we provide the notations deployed in this paper. 
Let $\mathcal{D}$ express the input passage with $m$ sentences, which is denoted as $\mathcal{D} = \left\{ \boldsymbol{S}_1,  \boldsymbol{S}_2,\dots,  \boldsymbol{S}_m\right\}$.
Each sentence $\boldsymbol{S}_i$ is composed of $n_i$ tokens, i.e., $ \boldsymbol{S}_i = \left\{t_i^1,t_i^2,\dots,t_i^{n_i}\right\}$. In addition, the set of entity tokens in $ \boldsymbol{S}_i$ is formulated by $ \boldsymbol{E}_i = \left\{e_i^1,e_i^2,\dots,e_i^{| \boldsymbol{E}_i|}\right\}$, where $|\boldsymbol{E}_i|$ indicates the number of entities in the $i$-th sentence.

\subsection{Token-level Uncertainty}
Generally, the conditional probability of a token output by LLMs reflects its likelihood in the context, which can be adapted to measure the uncertainty. 
Previous researches mainly focus on using the negative log probability or entropy-based methods for uncertainty estimation \citep{DBLP:journals/corr/abs-2307-10236}. 
In this paper, we integrate two statistical indicators, namely the maximum and variance of the probability distributions. 
Moreover, hallucination tends to accumulate with the increasing sequence length as demonstrated in previous studies \citep{DBLP:journals/corr/abs-2307-03987,DBLP:conf/emnlp/NaszadiMM23,DBLP:journals/corr/abs-2403-00896}, thus we further devise a sequence decay term that explicitly models the absolute position of the token in the passage. 
Specifically, the token-level uncertainty in the $j$-th position of $i$-th sentence can be measured as:
\begin{equation}
\label{eq:token}
\mathcal{U}(t_i^j) = \frac{1}{\max(\mathcal{C}_i^j)+ \sigma^2(\mathcal{C}_i^j)} \underbrace{(1+e^{\frac{len(\boldsymbol{S}_{1:i-1}) + j}{len(\mathcal{D})}-1})}_{\text{sequence decay term}}
\end{equation}
where $\mathcal{C}_i^j$ signifies the top-$k$ probabilities of a candidate token set that could probably appear in the current position based on LLMs, which is formulated as:
\begin{equation}
\mathcal{C}_i^j = sorted(P^{\mathcal{V}}_{ij})\left[-k:\right] \nonumber
\end{equation}
where $P^{\mathcal{V}}_{ij}$ expresses the list of all probabilities for the vocabulary at the $j$-th position of $i$-th sentence. $\max()$ and $\sigma^2()$ represent the maximum and variance functions separately. 
If the values of maximum and variance are high, the model will be more confident about its output. 
The second term is a sequence decay we designed to increase the uncertainty of tokens when the length of the generated sequence grows. 
$len(\mathcal{D})$ is the total number of tokens in the entire passage, and $len(\boldsymbol{S}_{1:i-1}) + j$ shows the position of the current token in the passage. 




\subsection{Sentence-level Uncertainty}
Previous works \citep{DBLP:conf/naacl/PagnoniBT21,DBLP:conf/emnlp/KryscinskiMXS20} illustrate that a major of hallucination in text generation is induced by the entity errors, such as false relations between two entities, inconsistent mentions in the context or basic factual errors, etc. 
This corresponds to our intuition that humans usually pay more attention to the salient information such as the keywords or entities for verification of the generated results. 
Therefore, recent researches turn to investigate the uncertainty of informative and important entities for hallucination detection. 
However, the complex dependencies over the entities are not well studied. 
In this paper, we explore the relations in the constructed semantic graph for uncertainty propagation and hallucination estimation. 



\paragraph{Relation-based Uncertainty Propagation.}
Previous findings reveal that each token influences the surrounding context \citep{DBLP:conf/sigir/ChenHHHA17}, thus the hallucination would probably propagate across the generated text. 
\citet{DBLP:conf/emnlp/ZhangQGDZZZWF23} presents a hallucination propagation method that propagates the uncertainty score of preceding entity tokens to the current one. 
Whereas, this method roughly uses all the preceding entities, while ignoring their potential dependency relations with the current entity, which is inclined to overestimate the uncertainties by our preliminary studies. 
In this paper, we present a relation-based uncertainty propagation method and assume that the subject entity propagates its uncertainty to the object entity based on the predicate or relation in the semantic graph. 
Moreover, we devise a penalty factor based on the relation intensity to alleviate the uncertainty overestimation problem. 

To be specific, given an object entity $o$, we first search the entities that have semantic relations with $o$ from the semantic graph and obtain a set of triples as $\mathcal{T}_o = \{(s', v', o) | (s', v', o) \in \mathcal{T}_i\}$.
$\mathcal{T}_i$ is the set of triples in semantic graph of sentence $i$.
Intuitively, the subject entities are not equally important to the object entity, thus we leverage their attention scores as the weights for uncertainty propagation. 
To alleviate the overestimation problem, we additionally incorporate a relation intensity-based penalty factor for propagation. 
The final uncertainty of an object entity is formulated as:
\begin{equation}
\label{eq:propa}
\mathcal{U}_p(o) =  \sum_{(s', v', o) \in \mathcal{T}_o} \frac{att(s', o)}{\mathcal{I}_o} * \mathcal{U}(s')
\end{equation}
where $att(,)$ signifies the attention score between two tokens, $\mathcal{I}_o$ is a penalty factor that computes the relation intensity of all entities that have relations with the object $o$, which can be measured as follows:
\begin{equation}
\label{eq:pathave}
\mathcal{I}_o = \frac{1}{|\mathcal{T}_o|} \sum_{(s', v', o) \in \mathcal{T}_o}\frac{att(s', v') + att(v', o)}{2}
\end{equation}
In general, high relation intensities usually indicate high factuality-confidence, thus the propagated uncertainties should be penalized. 

\paragraph{Entity Uncertainty.}
For an entity $e_i^j$, the uncertainty score consists of the self-uncertainty (Formula~\ref{eq:token}) and the propagated uncertainty (Formula~\ref{eq:propa}). 
The entity-based uncertainty of sentence $S_i$ can be calculated by averaging the uncertainties of all entities in the sentence:
\begin{equation}
\label{eq:entityu}
\mathcal{U}_{E}(i) = \frac{1}{|\boldsymbol{E}_i|} \sum\nolimits_{e_i^j \in \boldsymbol{E}_i} [\mathcal{U}(e_i^j) + \beta \mathcal{U}_p(e_i^j)]
\end{equation}
where $\mathcal{U}(e_i^j)$ and $\mathcal{U}_p(e_i^j)$ show the self-uncertainty and propagated uncertainty respectively, and $\beta$ is a hyper-parameter to balance these two uncertainties.

\paragraph{Global Uncertainty.}
In addition to the entities, there are also many general tokens in the sentence. 
To capture the global information of the sentence, we also
consider the uncertainties of all tokens (both entities and general tokens) in the sentence, and utilize the quantile approach to measure the global uncertainty, which is effective in capturing the global statistics in distributions \cite{DBLP:journals/corr/abs-2404-10136}:
\begin{equation}
\label{eq:globalu}
\mathcal{U}_{G}(i) = \text{qua}_{\alpha}(\mathcal{U}(t_i^1:t_i^{n_i}))
\end{equation}
where $\text{qua}_{\alpha}(\mathcal{U}(t_i^1:t_i^{n_i}))$ is the $\alpha$-quantile of the uncertainties of all tokens in sentence $\boldsymbol{S}_i$.


The uncertainty of the $i$-th sentence is the interpolation sum of the entity-based uncertainty and the global uncertainty:
\begin{equation}
\label{eq:senu}
\mathcal{U}_{s}(i) =  \lambda \mathcal{U}_{E}(i)+(1-\lambda)\mathcal{U}_{G}(i)
\end{equation}
where $\lambda$ is an interpolation weight.

\subsection{Passage-level Uncertainty}
Previous methods usually estimate the average uncertainty of all sentences for passage-level uncertainty.
However, the intricate relations among the sentences are neglected, which could affect the detection of hallucination where two sentences contradict each other despite each sentence having low uncertainty. 
For example, the first sentence in a passage is \textit{`Thomas was born in 1972.'} and the fourth sentence is \textit{`He raced until 1968.'}, which are contradictory in the passage.
In this paper, we present a graph-based uncertainty calibration method that incorporates the relations of the sentence-centered sub-graph for uncertainty calibration. 
The calibrated uncertainties of all sentences are averaged as the passage-level uncertainty.



\paragraph{Graph-based Uncertainty Calibration.}
Intuitively, if a sentence contradicts all the neighbor sentences in the semantic graph, it will probably have inconsistency or conflicts in the context, which is prone to the hallucination problem. 
Thus, the uncertainty score should be increased. 
Motivated by this intuition, we present a graph-based uncertainty calibration method. 
First, we search the neighbor nodes for each sentence from the semantic graph. Then, we calculate the contradictory score for each connected sentence pair with a NLI model, namely DeBERTa-v3-Large \citep{DBLP:conf/iclr/HeGC23}, which is widely applied for natural language processing tasks. 
Finally, we incorporate the uncertainty of each sentence with the neighbor contradictory scores for passage-level uncertainty computation:
\begin{equation}
\label{eq:passu}
\mathcal{U}_p = \frac{1}{\sum_{i=1}^m |\mathcal{N}(i)|} \sum_{i=1}^m \sum_{j \in \mathcal{N}(i)} \mathcal{U}_{s}(i)*\text{NLI}(con |\boldsymbol{S}_j,\boldsymbol{S}_i)
\end{equation}
where $\mathcal{N}(i)$ reflects the neighbors of the $i$-th sentence in the graph, $\text{NLI}(con|,)$ is the contradiction probability between two sentences via the NLI model.


\section{Experimental Setup}
\paragraph{Datasets}
 We conduct extensive experiments on two datasets for hallucination detection. 
One is currently the latest and most widely used dataset WikiBio.
To verify the effectiveness and generalization of our method, we also construct a Chinese dataset NoteSum, which can help boost research in this area. 
\textbf{WikiBio} \citep{DBLP:conf/emnlp/ManakulLG23} is a dataset derived from Wikipedia biographies. 
WikiBio applies the names from Wikipedia as the topics and generates corresponding biographies using GPT-3 \citep{DBLP:journals/mima/FloridiC20}.
Each sentence is annotated with one of the following labels: Factual (hallucination score: 0), NonFact* (0.5), and NonFact (1), which indicates a sentence with no hallucination, with factual errors, and is irrelevant to the topic respectively. 
The entire passage also has a human-labeled hallucination score as the ground truth.
\textbf{NoteSum} is an industrial Chinese dataset. 
The company first collects users' long text notes on various daily topics with numerous entities. 
We cooperate with the company and create shorter summaries from these long notes by LLMs for research. 
The private information of users is removed. 
It consists of both factuality and faithfulness hallucination as WikiBio.
We also adopt the same annotation guideline with WikiBio.
The statistics of the datasets are shown in Table~\ref{tab1}.

\begin{table}[t]
\centering
\renewcommand{\arraystretch}{1.0} 
 \setlength{\tabcolsep}{1.5mm}{
\begin{tabular}{|l|cc|}
\toprule
    & \textbf{WikiBio} & \textbf{NoteSum}\\
\midrule
Language & English & Chinese \\
\# Passages & 238 & 200\\
\# Sentences & 1908 & 1004 \\
\# Words/Sentence &  17.49  & 33.38\\
\# Sentences/Passage &  8.02  & 5.02 \\
 Halu Rate (\%)  &  72.95  & 65.27 \\
 Fact Halu Rate (\%)  & 33.07  & 27.94 \\
 Faith Halu Rate (\%)  & 39.88  & 37.33 \\ 
\bottomrule
\end{tabular}}
\caption{Statistics of WikiBio and NoteSum. `Fact Halu Rate (\%)' and `Faith Halu Rate (\%)' demonstrate the proportion of sentences with factuality and faithfulness hallucination.}
\label{tab1}
\end{table}

\paragraph{Evaluation Metrics}
 For fair comparison, we apply the evaluation metrics used in previous researches \citep{DBLP:conf/emnlp/ManakulLG23,DBLP:conf/emnlp/ZhangQGDZZZWF23}. Specifically, the area under curves (AUC) \citep{DBLP:journals/pr/Bradley97} are used to measure the performance of sentence-level hallucination detection.
 To evaluate the agreement between the passage-level hallucination score and human judgment, we employ the Pearson correlation coefficient \citep{cohen2009pearson} and the Spearman correlation coefficient \citep{sedgwick2014spearman} to estimate the degree of consistency.


\begin{table*}[t]
\renewcommand{\arraystretch}{1.0} 
 \setlength{\tabcolsep}{1.5mm}{
\begin{tabular}{|lcllclcllcl|}
\toprule
\multicolumn{1}{|c|}{}     & \multicolumn{5}{c|}{\textbf{WikiBio}}  & \multicolumn{5}{c|}{\textbf{NoteSum}} \\ 
\multicolumn{1}{|c|}{\multirow{3}{*}{Methods}} & \multicolumn{3}{c}{sentence-level} & \multicolumn{2}{c|}{passage-level}     & \multicolumn{3}{c}{sentence-level} & \multicolumn{2}{c|}{passage-level} \\ 
\multicolumn{1}{|c|}{}   & \multicolumn{1}{c}{NonFact} & \multicolumn{1}{c}{NonFact*} & \multicolumn{1}{c}{Factual} & \multicolumn{1}{c}{Pearson} & \multicolumn{1}{c|}{Spearman} & \multicolumn{1}{c}{NonFact} & \multicolumn{1}{c}{NonFact*} & \multicolumn{1}{c}{Factual} & \multicolumn{1}{c}{Pearson} & \multicolumn{1}{c|}{Spearman} \\ 
\midrule
\multicolumn{11}{|l|}{GPT-3 Uncertainties}  \\ 
\midrule
\multicolumn{1}{|l|}{Avg(-log$p$)}   & \multicolumn{1}{c}{83.21}  & \multicolumn{1}{c}{38.89}      & \multicolumn{1}{c}{53.97}   & \multicolumn{1}{c}{57.04}    & \multicolumn{1}{c|}{53.93}    & \multicolumn{1}{c}{80.11}    & \multicolumn{1}{c}{43.69}    & \multicolumn{1}{c}{35.29}        & \multicolumn{1}{c}{39.61}    &   \multicolumn{1}{c|}{31.55} \\ 
\multicolumn{1}{|l|}{Avg($\mathcal{H}$)}   & \multicolumn{1}{c}{80.73}  & \multicolumn{1}{c}{37.09}      & \multicolumn{1}{c}{52.07}   & \multicolumn{1}{c}{55.52}    & \multicolumn{1}{c|}{50.87}    & \multicolumn{1}{c}{80.08}    & \multicolumn{1}{c}{43.95}    & \multicolumn{1}{c}{38.04}        & \multicolumn{1}{c}{40.36}    &   \multicolumn{1}{c|}{33.25} \\ 
\multicolumn{1}{|l|}{Max(-log$p$)}   & \multicolumn{1}{c}{87.51}  & \multicolumn{1}{c}{35.88}      & \multicolumn{1}{c}{50.46}   & \multicolumn{1}{c}{57.83}    & \multicolumn{1}{c|}{55.69}    & \multicolumn{1}{c}{79.86}    & \multicolumn{1}{c}{40.17}    & \multicolumn{1}{c}{36.70}        & \multicolumn{1}{c}{38.13}    &   \multicolumn{1}{c|}{34.75} \\ 
\multicolumn{1}{|l|}{Max($\mathcal{H}$)}   & \multicolumn{1}{c}{85.75}  & \multicolumn{1}{c}{32.43}      & \multicolumn{1}{c}{50.27}   & \multicolumn{1}{c}{52.48}    & \multicolumn{1}{c|}{49.55}    & \multicolumn{1}{c}{81.02}    & \multicolumn{1}{c}{47.33}    & \multicolumn{1}{c}{39.03}        & \multicolumn{1}{c}{42.88}    &   \multicolumn{1}{c|}{37.24} \\ 
\midrule
\multicolumn{11}{|l|}{SelfCheckGPT (gpt-3.5-turbo)}  \\ 
\midrule
\multicolumn{1}{|l|}{BertScore}   & \multicolumn{1}{c}{81.96}  & \multicolumn{1}{c}{45.96}      & \multicolumn{1}{c}{44.23}   & \multicolumn{1}{c}{58.18}    & \multicolumn{1}{c|}{55.90}    & \multicolumn{1}{c}{76.44}    & \multicolumn{1}{c}{39.69}    & \multicolumn{1}{c}{36.89}        & \multicolumn{1}{c}{25.91}    &   \multicolumn{1}{c|}{21.24} \\ 
\multicolumn{1}{|l|}{QA}   & \multicolumn{1}{c}{84.26}  & \multicolumn{1}{c}{40.06}      & \multicolumn{1}{c}{48.14}   & \multicolumn{1}{c}{61.07}    & \multicolumn{1}{c|}{59.29}    & \multicolumn{1}{c}{79.69}    & \multicolumn{1}{c}{45.30}    & \multicolumn{1}{c}{39.32}        & \multicolumn{1}{c}{41.07}    &   \multicolumn{1}{c|}{36.54} \\ 
\multicolumn{1}{|l|}{Unigram (max)}   & \multicolumn{1}{c}{85.63}  & \multicolumn{1}{c}{41.04}      & \multicolumn{1}{c}{58.47}   & \multicolumn{1}{c}{64.71}    & \multicolumn{1}{c|}{64.91}    & \multicolumn{1}{c}{79.48}    & \multicolumn{1}{c}{43.88}    & \multicolumn{1}{c}{36.15}        & \multicolumn{1}{c}{38.80}    &   \multicolumn{1}{c|}{33.35} \\ 
\multicolumn{1}{|l|}{Combi}   & \multicolumn{1}{c}{87.33}  & \multicolumn{1}{c}{44.37}      & \multicolumn{1}{c}{61.83}   & \multicolumn{1}{c}{69.05}    & \multicolumn{1}{c|}{67.77}    & \multicolumn{1}{c}{82.38}    & \multicolumn{1}{c}{\underline{53.19}}    & \multicolumn{1}{c}{40.17}        & \multicolumn{1}{c}{47.79}    &   \multicolumn{1}{c|}{41.27} \\ 
\midrule
\multicolumn{11}{|l|}{FOCUS}  \\ 
\midrule
\multicolumn{1}{|l|}{LLaMA-13B}   & \multicolumn{1}{c}{87.90}  & \multicolumn{1}{c}{43.84}      & \multicolumn{1}{c}{62.46}   & \multicolumn{1}{c}{70.62}    & \multicolumn{1}{c|}{63.03}    & \multicolumn{1}{c}{81.11}    & \multicolumn{1}{c}{49.98}    & \multicolumn{1}{c}{38.88}        & \multicolumn{1}{c}{38.17}    &   \multicolumn{1}{c|}{38.31} \\
\multicolumn{1}{|l|}{LLaMA-30B}   & \multicolumn{1}{c}{89.79}  & \multicolumn{1}{c}{48.80}      & \multicolumn{1}{c}{\underline{65.69}}   & \multicolumn{1}{c}{\underline{77.15}}    & \multicolumn{1}{c|}{\underline{73.24}}    & \multicolumn{1}{c}{82.17}    & \multicolumn{1}{c}{43.12}    & \multicolumn{1}{c}{49.85}        & \multicolumn{1}{c}{37.37}    &   \multicolumn{1}{c|}{40.09} \\
\midrule
\multicolumn{11}{|l|}{\textbf{OURS}}  \\ 
\midrule
\multicolumn{1}{|l|}{LLaMA-13B}   & \multicolumn{1}{c}{\underline{90.14}}  & \multicolumn{1}{c}{\textbf{61.65}}      & \multicolumn{1}{c}{64.82}   & \multicolumn{1}{c}{72.11}    & \multicolumn{1}{c|}{64.35}    & \multicolumn{1}{c}{\underline{85.06}}    & \multicolumn{1}{c}{50.70}    & \multicolumn{1}{c}{\underline{53.03}}        & \multicolumn{1}{c}{\textbf{55.62}}    &   \multicolumn{1}{c|}{\underline{60.81}} \\
\multicolumn{1}{|l|}{LLaMA-30B}   & \multicolumn{1}{c}{\textbf{90.93}}  & \multicolumn{1}{c}{\underline{61.16}}      & \multicolumn{1}{c}{\textbf{65.70}}   & \multicolumn{1}{c}{\textbf{77.60}}    & \multicolumn{1}{c|}{\textbf{74.44}}    & \multicolumn{1}{c}{\textbf{87.95}}    & \multicolumn{1}{c}{\textbf{54.42}}    & \multicolumn{1}{c}{\textbf{61.51}}        & \multicolumn{1}{c}{\underline{54.77}}    &   \multicolumn{1}{c|}{\textbf{61.05}} \\
\midrule
\multicolumn{1}{|c|}{$\Delta$} &\multicolumn{1}{c}{+1.14} &\multicolumn{1}{c}{+12.85} &\multicolumn{1}{c}{+0.01} &\multicolumn{1}{c}{+0.45} &\multicolumn{1}{c|}{+1.20} &\multicolumn{1}{c}{+5.57} &\multicolumn{1}{c}{+1.23} &\multicolumn{1}{c}{+11.66} &\multicolumn{1}{c}{+7.83} &\multicolumn{1}{c|}{+19.78} \\
\bottomrule
\end{tabular}}
\caption{Comparison results of our approach and the recent hallucination detection methods. The best results are in \textbf{bold} and the second best is marked with \underline{underline}. $\Delta$ indicates our maximum improvements over the best baselines.}
\label{tab2}
\end{table*}

\paragraph{Baselines}
We compare our approach with the recent advanced baselines:
1) \textbf{GPT-3 Uncertainties} method uses the GPT-3 model to output the probability of each token, and then
various uncertainty metrics are calculated as \citet{DBLP:conf/emnlp/ManakulLG23} do for hallucination detection.
2) \textbf{SelfCheckGPT} \citep{DBLP:conf/emnlp/ManakulLG23} is the recent sampling-based method that relies on frequent sampling from LLMs for consistency checking. The gpt-3.5-turbo model is used and four methods are applied to measure the consistency, namely BertScore, QA, Unigram, and their combination.
3) \textbf{FOCUS} \citep{DBLP:conf/emnlp/ZhangQGDZZZWF23} is currently the outstanding uncertainty-based detection method. 
We leverage the LLaMA-13B and LLaMA-30B as the backbones.

\paragraph{Implementation Details}
we utilize a transition-based AMR parser \citep{xu-etal-2023-learning-friend} to construct an AMR graph for each sentence. Then, we perform coreference resolution and entity linking by spaCy to link sentence-level AMR graphs together to obtain a passage-level graph for each passage. The DeBERTa-v3-Large \citep{DBLP:conf/iclr/HeGC23} NLI model is used to calculate the contradiction probability in Formula~\ref{eq:passu}. 
We experiment with the LLaMA-13B and LLaMA-30B models to obtain the probability of each token.
The hyper-parameters $\alpha$, $\beta$, $\lambda$, and $k$ are set to 0.8, 0.65, 0.7, and 3 respectively.

\section{Results and Analyses}
\subsection{Main Results}
Table~\ref{tab2} shows the performance of our approach and the state-of-the-art baselines.
We have the following observation.
\textbf{First,} we achieve the best performance on both sentence-level and passage-level hallucination detection regarding all evaluation metrics. 
In particular, we gain a maximum improvement of 19.78\% over the best baseline in passage-level hallucination detection.
\textbf{Second,} compared with FOCUS that propagates the uncertainties of all preceding focused tokens to the subsequent one, our approach yields significant improvements especially for the NonFact* and Factual types that have moderate and no hallucination respectively, indicating the effectiveness of our relation-based uncertainty propagation to help alleviate the overestimation problem.
\textbf{Third,} our approach exhibits good cross-domain and cross-language generalization. 
It not only performs well on the English biography dataset WikiBio, but also reflects significant improvements on the Chinese note summary dataset NoteSum.

\subsection{Ablation Studies}
We conduct ablation studies on WikiBio with LLaMA-30B from three dimensions: token, sentence, and passage.
Experimental results are shown in Table~\ref{tab3}.
For each row, one setting is removed while keeping the other settings unchanged.

\begin{table}[t]
\renewcommand{\arraystretch}{1.0} 
 \setlength{\tabcolsep}{1.5mm}{
\begin{tabular}{|l|ccccc|}
\toprule
& \multicolumn{3}{c}{sentence-level} & \multicolumn{2}{c|}{passage-level}\\
      & NonFact & NonFact* & Fact &Pear. & Spear. \\
      \midrule
 Ours & 90.93 & 61.16 & 65.70 & 77.60 & 74.44 \\
    \midrule
 - max & 86.48 & 64.86  & 63.52 & 23.32 & 38.57  \\
 - var & 90.17 & 50.94 & 64.82  & 75.60 & 72.36 \\
 - decay & 89.01 & 43.57 & 63.48 & 70.19 & 66.49 \\
 \midrule
 - entity & 88.31 & 43.06 & 63.10 & 65.81 & 60.34\\
 - global & 88.75 & 43.88 & 65.19 & 70.36 & 65.49 \\
 \midrule
 - graph & - & - & - & 75.89 & 72.20 \\
 
\bottomrule
\end{tabular}}
\caption{Results of ablation studies on WikiBio. `- max', `- var' and `- decay' mean removing the maximum, variance and decay term from Formula~\ref{eq:token}.
`- entity' and `- global' reveal removing the entity and global uncertainty respectively from Formula~\ref{eq:senu}. `- graph' indicates not including the contradiction probability of the neighbors in the graph, i.e., averaging the uncertainties of all sentences in Formula~\ref{eq:passu}. }
\label{tab3}
\end{table}

We have the following observations: 
(1) By removing each element from Formula~\ref{eq:token} respectively, the performance decreases significantly in most cases, which signifies the effectiveness of the maximum, variance, and decay term for modeling the token-level uncertainty.
(2) The performance with the passage-level metrics drops more significantly with the setting of `- max', manifesting that the maximum probability can better capture the key features for hallucination detection, while other terms can help further refine the uncertainty.
(3) Both the entity uncertainty computed by relation-based propagation and the global uncertainty are important to sentence-level detection. 
In addition, entity uncertainty is more effective than global uncertainty for passage-level detection.
(4) By excluding the contradiction relations of the neighbor sentences in the semantic graph, the performance of passage-level hallucination detection significantly drops by about 2 points, which further verifies the effectiveness of our graph-based uncertainty calibration for detecting hallucination over the passage.

\begin{figure}[t]
\centering
\includegraphics[width=0.44\textwidth,height=0.22\textheight]{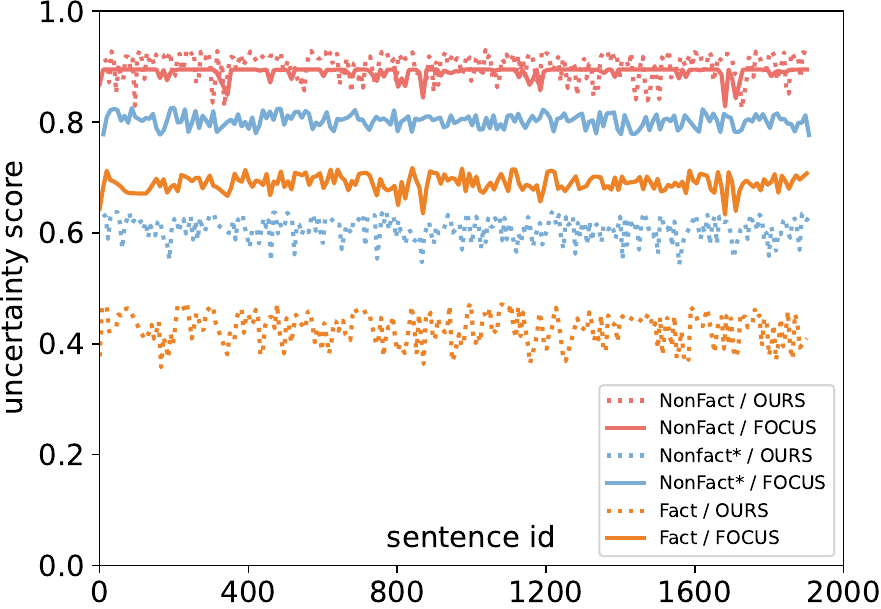} %
\caption{The uncertainty scores of three types of samples calculated with FOCUS and ours.}
\label{fig3}
\end{figure}

\begin{figure}[t]
\centering
\includegraphics[width=0.4\textwidth,height=0.18\textheight]{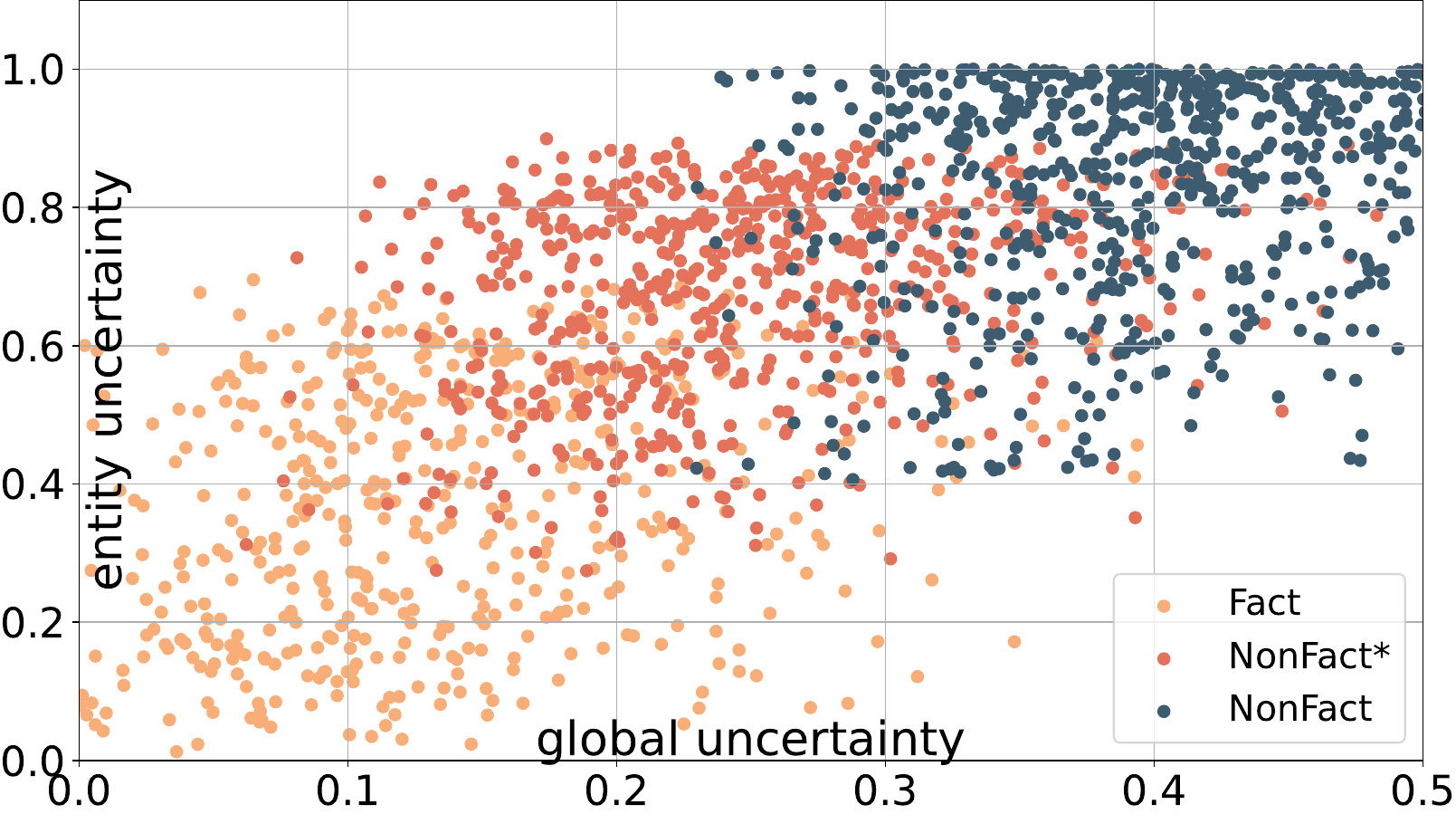} %
\caption{Visualization of the entity uncertainty and global uncertainty for three types of samples.}
\label{fig40}
\end{figure}

\subsection{Further Analyses}

\paragraph{Effect of Relation-based Uncertainty Propagation.}
To further investigate the effectiveness of our relation-based uncertainty propagation method, we compare with the baseline FOCUS \citep{DBLP:conf/emnlp/ZhangQGDZZZWF23} that propagates the uncertainties of all preceding keywords to the subsequent one. 
The results are shown in Figure~\ref{fig3}, illustrating the uncertainty scores of three types of samples from WikiBio measured by FOCUS and ours respectively. 
We can observe that both of the two methods yield high uncertainty scores for the samples with NonFact (ground truth score = 1), which can help well identify the severe hallucination.
It is also notable that the FOCUS method tends to overestimate the uncertainties for the samples with NonFact* (ground truth score = 0.5) and Fact (ground truth score = 0). 
There is a large gap between the estimated uncertainties and the ground truth. 
Moreover, the uncertainties of the three types calculated by FOCUS are very close, making it difficult to identify hallucination in different degrees. 
In contrast, our approach effectively diminishes the uncertainties for samples with NonFact* and Fact, which further verifies the effectiveness of our relation-based uncertainty propagation in alleviating the overestimation problem.


\paragraph{Visualization of Entity and Global Uncertainty.}
To examine the effect of entity and global uncertainty for sentence-level hallucination detection, scores of the two uncertainties are visualized for three types of samples from WikiBio in Figure~\ref{fig40}. 
We observe that with the increased degree of hallucinations (Fact $\rightarrow$ NonFact* $\rightarrow$ NonFact), both types of uncertainty scores increase. Moreover, there are fewer overlaps in the three types of samples based on entity uncertainty and global uncertainty. 
In other words, the three types of samples can be well distinguished by the two uncertainties. 
All these observations demonstrate the effectiveness of our entity and global uncertainty. 

\begin{figure}[t]
\centering
\includegraphics[width=0.3\textwidth,height=0.18\textheight]{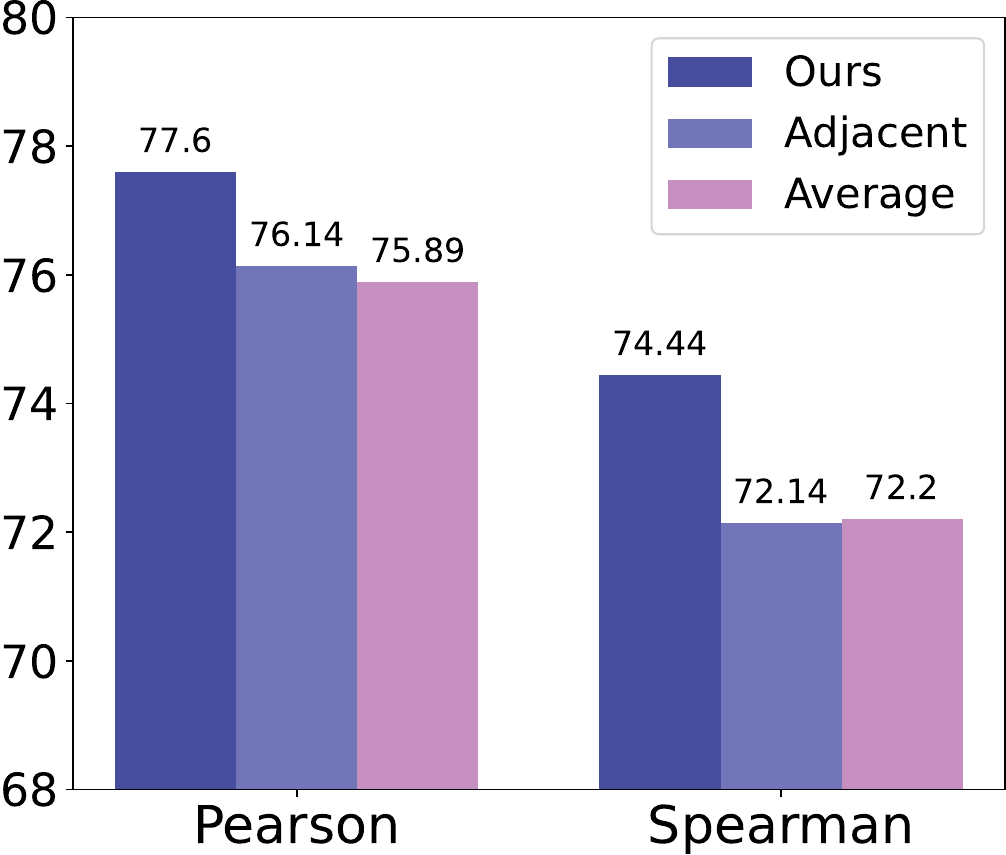} %
\caption{The Pearson and Spearman metrics of ours and the compared methods for passage-level uncertainty calculation.}
\label{fig5}
\end{figure}

\paragraph{Effect of Graph-based Uncertainty Calibration.}
To verify the effectiveness of our graph-based uncertainty calibration, we compare it with other two methods, namely Adjacent and Average. 
The Adjacent method merely incorporates the relations between the current sentence and the previous as well as the next sentence for uncertainty calculation, while the Average method simply measures the average uncertainties of all sentences. 
The results of the two methods and ours are shown in Figure~\ref{fig5}. 
Our method is observed to outperform Adjacent and Average in terms of Pearson and Spearman correlations, indicating the effectiveness of using the semantic graph to model the long-range sentence relations for passage-level hallucination detection. 
In addition, the performance of Adjacent and Average is close, indicating the limits of merely considering the adjacent sentences. 



\section{Conclusions}
In this paper, we propose a method to enhance uncertainty modeling with semantic graph for hallucination detection. Extensive experiments verify the effectiveness of each component of our approach. 
In particular, our approach consistently outperforms the state-of-the-art baselines in both sentence-level and passage-level hallucination detection, by incorporating the semantic relations among entities and sentences into the uncertainty calculation framework. 
It is also interesting to find that our relation-based uncertainty propagation method can help effectively alleviate the uncertainty overestimation problem and our graph-based uncertainty calibration method can capture long-range relations. 
In the future, we will explore integrating the existing knowledge graph with AMR graphs for fact-checking and hallucination detection. 

\section{Ethics Statement}
Our WikiBio dataset is publicly used in the field of natural language processing. The NoteSum dataset, on the other hand, is an internal private dataset of Xiaohongshu, and its construction, annotation, and review are all handled by Xiaohongshu's own employees. The method presented in this paper is original to the authors and does not involve any ethical issues.

\section{Acknowledgments}
Thanks to all collaborators and reviewers for their efforts.This research is funded by the National Science and Technology Major Project (No. 2021ZD0114002), the National Nature Science Foundation of China (No. 62477010), the Science and Technology Commission of Shanghai Municipality Grant (No. 22511105901, No. 21511100402), Shanghai Science and Technology Innovation Action Plan (No. 24YF2710100), and Shanghai Special Project to Promote High-quality Industrial Development (No. RZ-CYAI-01-24-0288).

\bibliography{aaai25}

\newpage
\clearpage
\appendix
\setcounter{secnumdepth}{2}
\section{Appendices}

\subsection{Details for Semantic Paths}
\label{path}
First, we use the AMR Python library `amrlib' to parse each sentence semantically. 
Then, based on the output semantic sentence-graph structures, we extract all dependency-related triples using handwritten rules.
Next, we apply the spaCy library to tag the word class of the three elements in all the triples. 
We ensure that all the head and tail entities of all triples are noun entities with pos\_tag (`NOUN', `NUM', `PROPN') or with ner\_type (`PERSON', `DATE', `ORG', `GPE', `NORP', `ORDINAL', `PRODUCT', `CARDINAL', `LOC', `FAC', `EVENT',
`WORK\_OF\_ART', `LAW', `LANGUAGE', `TIME', `PERCENT', `MONEY', `QUANTITY'), while the relation entities are verb words (`VERB'). 
Finally, we manually review all the extracted triples to obtain more accurate results.

\subsection{The Construction Details of NoteSum Dataset}
\label{NoteSum}
\paragraph{Overall Information}
NoteSum is a dataset of Chinese note summaries.
One of our company's key businesses is producing electronic notes on mobile devices to facilitate user access.
Due to certain special business needs, we need to generate shorter summaries of these long notes.
In this process, the summaries we generate must not contain factual errors and must accurately reflect the content of the original notes.
To achieve this goal, company staff need to conduct strict reviews of the generated note summaries.

With the advent of the large language model era, all note summaries are generated by the LLMs trained by the company.
Therefore, the task of reviewing note summaries can be seen as a hallucination detection task for LLMs. 
Factual errors correspond to factuality hallucination, while checking the alignment between the summary and the original text can be considered as faithfulness hallucination detection.

\paragraph{Construction Process}
We first manually select 200 notes about science or introduction that contained a relatively large number of entities.
Subsequently, we employ the LLM trained by our company to generate summaries for these 200 samples with the prompt below.
Finally, we invite two experts from the product department to annotate all the samples, following the annotation guidelines of the WikiBio dataset: 
Factual (0), NonFact* (0.5), or NonFact (1), representing no hallucination, a sentence with factual errors, and a sentence that is irrelevant to the origin note, respectively.
The entire passage also has a hallucination score as the ground truth. 
The agreement score of their annotations is 0.76.\\
-----------------------------------------------------------------------

\textbf{prompt for summary generation}:

Please generate a Chinese summary for the following notes within 3 to 6 sentences in Chinese: [NOTE CONTENT] (in Chinese).\\
-----------------------------------------------------------------------

\subsection{The Supplementary for Metrics}
\label{metrics}
We use AUC as the evaluation metric for the three sentence-level labels.
However, the basic setup for these three types of labels is different, following the same computing approach of \citet{DBLP:conf/emnlp/ManakulLG23,DBLP:conf/emnlp/ZhangQGDZZZWF23}.

\paragraph{NonFact} We set the ground truth of sentence samples with labels `NonFact' and `NonFact*' to 1, and the sentence samples with the label `Fact' to 0.
It focuses on the method’s ability to judge hallucination samples with both factulaity hallucination and faithfulness hallucination.

\paragraph{NonFact*} We set the ground truth of sentence samples with the label `NonFact' to 1, and the sentence samples with labels `NonFact*' and `Fact' to 0.
It focuses on the method’s ability to judge fully hallucinated samples with faithfulness hallucination.

\paragraph{Factual} Normal computational approach as usual.

\subsection{Detailed Information for Baselines}
\label{baselines}
\paragraph{ChatGPT3.5} is the most famous close-source LLM \citep{liu2023summary} trained by OpenAI. 
The API Key of the `text-davinci-003' version can output the probability of each token, which is now merged into the `gpt-3.5-turbo-instruct' version.
We apply the same uncertainty metrics, such as negative
log probability and entropy, with \citet{DBLP:conf/emnlp/ManakulLG23}.
We can calculate the uncertainty score for each sentence example and judge whether there is hallucination exists.

\paragraph{SelfCheckGPT} is the latest and popular black-box hallucination detection method \citep{DBLP:conf/emnlp/ManakulLG23}.
It is a kind of sampling-based hallucination detection method.
It rephrases the contents to be detected while ensuring the consistency of semantics by LLMs (ChatGPT3.5 version: gpt-3.5-turbo) with different temperatures. 
Furthermore, it calculates the consistency between the original and the rephrased contents using some sub-methods.
Here, we choose 4 sub-methods: BertScore, QA, Unigram and their combination.
(1) BerScore is used to compute the consistency of each rephrased sentence and the origin sentence.
(2) QA is to generate some questions and estimate the consistency between these questions. 
Here, we generate two questions for each sample in the NoteSum dataset.
(3) Unigram indicates approximating the closed-sourced LLMs’ token probabilities using the new language model.
(4) Combination means integrating normalized scores of the former three sub-methods.
To ensure a fair comparison, we do not use the prompt-based version.

\paragraph{FOCUS} \citep{DBLP:conf/emnlp/ZhangQGDZZZWF23} is an improved version of SelfCheckGPT. 
It is the most powerful uncertainty-based hallucination detection method.
It supposes that hallucination between entity tokens propagate based on high attention scores, thus adding a penalty. 
We select the well-performing LLaMA-13B and LLaMA-30B versions for the English dataset WikiBio.
When reproducing the results on the Chinese dataset NoteSum, we use the Chinese version of LLaMA with the same amount of parameters: Alpaca-Chinese-13B and Alpaca-Chinese-33B.
Additionally, the NLI model for this Chinese dataset is Deberta-Chinese-Large.
In a nutshell, we change all the models for the Chinese dataste NoteSum into the Chinese version.

\subsection{Prompts, Hyper-parameters and Projection Functions}
\label{implementation}
All the experiments are conducted on an A800 or A100 NVIDIA GPU with 80G graphic memory in a Linux environment. 
\paragraph{The Two Prompts}
We incorporate the text to be tested into the prompt and then input it into the LLM to obtain information such as the probability of each token, attention scores, and so on. 
The prompts for the two datasets are as follows:\\
-----------------------------------------------------------------------

\textbf{prompt for the WikiBio dataset}:

This is a passage from Wikipedia about [NAME OF A FAMOUS PERSON]: [PASSAGE TO BE DETECTED].\\
-----------------------------------------------------------------------\\
-----------------------------------------------------------------------

\textbf{prompt for the NoteSum dataset}:

This is a summary of the note. The note's body is [NOTE CONTENT], and its corresponding summary is [SUMMARY CONTENT]. (in Chinese)\\
-----------------------------------------------------------------------

\paragraph{The Hyper-parameters}
We adjust the following four parameters: $\alpha$, $\beta$, $\lambda$, and $k$ on the WikiBio dataset with LLaMA-30B, and the results are shown in Table~\ref{tab4}, Table~\ref{tab5}, Table~\ref{tab6} and Table~\ref{tab7}, respectively.
When adjusting the parameters, we ensure that the other parameters remain fixed.

According to the results in the tables, $\alpha$, $\beta$, $\lambda$, and $k$ are set as 0.8, 0.65, 0.7, and 3 respectively.

\begin{table}[t]
\renewcommand{\arraystretch}{1.0} 
 \setlength{\tabcolsep}{1.5mm}{
\begin{tabular}{|l|ccccc|}
\toprule
& \multicolumn{3}{c}{sentence-level} & \multicolumn{2}{c|}{passage-level}\\
      & NonFact & NonFact* & Fact &Pear. & Spear. \\
      \midrule
$\alpha = 0$   & 85.90 & 40.17 & 57.83 & 57.45 & 45.63 \\
$\alpha = 0.1$ & 86.94 & 41.32 & 61.07 & 61.22 & 51.35 \\
$\alpha = 0.2$ & 87.90 & 42.64 & 63.46 & 62.36 & 53.58 \\
$\alpha = 0.3$ & 87.89 & 42.65 & 63.32 & 61.14 & 50.19 \\
$\alpha = 0.4$ & 88.66 & 43.66 & 64.81 & 60.07 & 50.21 \\
$\alpha = 0.5$ & 88.63 & 43.61 & 64.71 & 60.56 & 50.52 \\
$\alpha = 0.6$ & 89.16 & 45.24 & 65.72 & 63.50 & 56.76 \\
$\alpha = 0.7$ & 90.23 & 56.41 & 66.25 & 76.26 & 72.03 \\
$\alpha = 0.8$ & 90.93 & 61.16 & 65.70 & 77.60 & 74.44  \\
$\alpha = 0.9$ & 90.58 & 66.53 & 64.02 & 76.67 & 73.84 \\
$\alpha = 1.0$ & 89.46 & 50.50 & 62.60  & 68.37 & 66.94 \\
\bottomrule
\end{tabular}}
\caption{Results about parameter $\alpha$. }
\label{tab4}
\end{table}

\begin{table}[t]
\renewcommand{\arraystretch}{1.0} 
 \setlength{\tabcolsep}{1.5mm}{
\begin{tabular}{|l|ccccc|}
\toprule
& \multicolumn{3}{c}{sentence-level} & \multicolumn{2}{c|}{passage-level}\\
      & NonFact & NonFact* & Fact &Pear. & Spear. \\
      \midrule
$\beta = 0.05$ & 83.42 & 37.73 & 49.32 & 45.31 & 29.76 \\
$\beta = 0.10$ & 84.51 & 38.89 & 53.39 & 48.29 & 34.41 \\
$\beta = 0.15$ & 85.65 & 39.96 & 57.14 & 53.13 & 41.30 \\
$\beta = 0.20$ & 86.77 & 41.09 & 60.12 & 58.17 & 48.88 \\
$\beta = 0.25$ & 87.78 & 42.48 & 62.58 & 62.43 & 55.55 \\
$\beta = 0.30$ & 88.55 & 43.47 & 64.25 & 66.13 & 60.71 \\
$\beta = 0.35$ & 89.08 & 45.04 & 65.39 & 70.26 & 66.27 \\
$\beta = 0.40$ & 89.68 & 48.14 & 66.08 & 73.38 & 69.25 \\
$\beta = 0.45$ & 90.22 & 52.15 & 66.32 & 76.01 & 72.31 \\
$\beta = 0.50$ & 90.85 & 56.37 & 66.27 & 77.45 & 73.42 \\
$\beta = 0.55$ & 90.79 & 60.67 & 65.67 & 77.43 & 74.01 \\
$\beta = 0.60$ & 90.91 & 63.48 & 65.15 & 77.68 & 73.95 \\
$\beta = 0.65$ & 90.93 & 61.16 & 65.70 & 77.60 & 74.44  \\
$\beta = 0.70$ & 90.56 & 67.52 & 63.89 & 76.69 & 74.20 \\
$\beta = 0.75$ & 90.20 & 67.03 & 63.46 & 75.41 & 73.88 \\
$\beta = 0.80$ & 89.67 & 67.52 & 62.58 & 74.41 & 74.05 \\
$\beta = 0.85$ & 89.16 & 66.89 & 62.13 & 73.25 & 73.88 \\
$\beta = 0.90$ & 88.53 & 66.27 & 61.27 & 72.39 & 73.61 \\
$\beta = 0.95$ & 88.09 & 66.08 & 60.91 & 71.45 & 73.71 \\
\bottomrule
\end{tabular}}
\caption{Results about parameter $\beta$. }
\label{tab5}
\end{table}

\begin{table}[t]
\renewcommand{\arraystretch}{1.0} 
 \setlength{\tabcolsep}{1.5mm}{
\begin{tabular}{|l|ccccc|}
\toprule
& \multicolumn{3}{c}{sentence-level} & \multicolumn{2}{c|}{passage-level}\\
      & NonFact & NonFact* & Fact &Pear. & Spear. \\
      \midrule
$\lambda = 0.1$ & 89.37 & 45.92 & 58.24 & 75.94 & 72.07 \\
$\lambda = 0.2$ & 89.68 & 47.22 & 59.68 & 76.44 & 72.10 \\
$\lambda = 0.3$ & 89.91 & 48.37 & 61.03 & 76.19 & 71.95 \\
$\lambda = 0.4$ & 90.01 & 49.31 & 62.17 & 77.03 & 71.96 \\
$\lambda = 0.5$ & 90.83 & 54.80 & 64.92 & 76.77 & 72.30 \\
$\lambda = 0.6$ & 90.90 & 65.34 & 64.54 & 76.96 & 73.38 \\
$\lambda = 0.7$ & 90.93 & 61.16 & 65.70 & 77.60 & 74.44  \\
$\lambda = 0.8$ & 90.95 & 60.68 & 65.34 & 77.18 & 73.36 \\
$\lambda= 0.9$  & 90.58 & 67.53 & 64.02 & 75.72 & 72.25 \\
\bottomrule
\end{tabular}}
\caption{Results about parameter $\lambda$. }
\label{tab6}
\end{table}

\begin{table}[t]
\renewcommand{\arraystretch}{1.0} 
 \setlength{\tabcolsep}{1.5mm}{
\begin{tabular}{|l|ccccc|}
\toprule
& \multicolumn{3}{c}{sentence-level} & \multicolumn{2}{c|}{passage-level}\\
      & NonFact & NonFact* & Fact &Pear. & Spear. \\
      \midrule
$k = 1$ & 90.17 & 50.94 & 64.82  & 75.60 & 72.36 \\
$k = 3$ & 90.93 & 61.16 & 65.70 & 77.60 & 74.44  \\
$k = 5$ & 90.33 & 53.15 & 66.25 & 76.21 & 72.57 \\
$k = 7$ & 89.93 & 49.33 & 66.19 & 74.43 & 70.60 \\
\bottomrule
\end{tabular}}
\caption{Results about parameter $k$. }
\label{tab7}
\end{table}

\paragraph{Projection Functions}
We treat the sentence-level and passage-level uncertainty scores as hallucination scores to determine whether hallucination exists. 
However, since both types of scores exceed 1, we need a projection function to convert them to a range between 0 and 1.

For each of the two uncertainties, we test on three projection functions: inverse, sigmoid, and logistic functions, on the WikiBio dataset with LLaMA-30B backbone.
The results are shown in Table~\ref{tab8}.

According to the results in the table, we choose a logistic function for sentences and an inverse function for passages.
\begin{table}[t]
\renewcommand{\arraystretch}{1.0} 
 \setlength{\tabcolsep}{1.5mm}{
\begin{tabular}{|l|ccccc|}
\toprule
& \multicolumn{3}{c}{sentence-level} & \multicolumn{2}{c|}{passage-level}\\
      & NonFact & NonFact* & Fact &Pear. & Spear. \\
      \midrule
s/inverse & 90.10 & 47.62 & 64.71 & - & - \\
s/sigmoid & 90.87 & 56.94 & 66.24  & - & - \\
s/logistic & 90.93 & 61.16 & 65.70 & - & -  \\
\midrule
p/inverse & - & - & - & 77.60 & 74.44  \\
p/sigmoid & - & - & - & 77.29 & 73.17 \\
p/logistic & - &-  & - & 75.80 & 73.61 \\
\bottomrule
\end{tabular}}
\caption{Results about the three kinds of projection functions of sentence-levle and passage-level uncertainty calculation. }
\label{tab8}
\end{table}

\subsection{Substitution Experiments}
\label{replacement}
To better illustrate the importance of each module in our method, we also conduct substitution experiments on the WikiBio dataset on LLaMA-30B backbone.
When we replace a specific module, we ensure that the other modules remain unchanged and consistent with the methods described in the paper.
The experimental results are in Table~\ref{tab9}.

The experimental results show that if some modules in our method are replaced with other computational methods, there will be a significant decrease in performance across five metrics, demonstrating the superiority of our method.
It effectively calculates uncertainty from the perspectives of tokens, sentences, and passages. 
These three types of uncertainty also greatly help us in detecting the presence of hallucination and determining its extent.

\begin{table}[t]
\renewcommand{\arraystretch}{1.0} 
 \setlength{\tabcolsep}{1.5mm}{
\begin{tabular}{|l|ccccc|}
\toprule
& \multicolumn{3}{c}{sentence-level} & \multicolumn{2}{c|}{passage-level}\\
      & NonFact & NonFact* & Fact &Pear. & Spear. \\
\midrule
 Ours & 90.93 & 61.16 & 65.70 & 77.60 & 74.44  \\
 \midrule
 w/ log & 82.71 & 36.98 & 46.92 & 45.60 & 29.66 \\
 \midrule
 w/ max(log) & 84.00 & 41.01 & 44.74 & 59.53 & 56.35 \\
 w/ ave(log) & 84.19 & 41.48 & 44.94 & 59.73 & 56.48 \\
 w/ max(ent) & 83.17 & 37.16 & 48.29 & 47.66 & 32.32 \\
 w/ ave(ent) & 84.09 & 41.16 & 44.86 & 59.63 & 56.45 \\
\midrule
 w/ adjacent & - & - & - & 76.14 & 72.14 \\
 w/ average & - & - & - & 75.89 & 72.20 \\
\bottomrule
\end{tabular}}
\caption{Results of substitution experiments. We conduct such experiments on three text granularity: token, sentence, and passage. `w/ log' means that we use the vanilla negative log probability as the token-level uncertainty. `w/ max(log)', `w/ ave(log)', `w/ max(ent)' and `ave(ent)' indicate the maximum of tokens' negative log probabilities, the average of tokens' negative log probabilities, the max of the entropy of tokens probabilities, and the average of the entropy of tokens probabilities as the sentence-level uncertainty separately. ` w/ adjacent' and ` w/ average' are calculating the degree of contradiction only within adjacent edges or merely
averaging the sentence scores as the passage score.}
\label{tab9}
\end{table}

\end{document}